\documentclass[journal]{IEEEtran}

\usepackage{xcolor}
\usepackage{amsmath}
\usepackage{graphicx}
\usepackage{booktabs}
\definecolor{chrxcolor}{RGB}{1, 140, 116}
\definecolor{firstteam}{RGB}{132, 94, 194}
\definecolor{secondteam}{RGB}{94, 194, 127}
\usepackage{amssymb}
\usepackage{cite}
\usepackage[colorlinks=true, citecolor=green, linkcolor=blue, urlcolor=cyan]{hyperref}

\ifCLASSINFOpdf
\else
\fi
\begin{document}
%
\title{Advancing All-Weather Building Damage Mapping to the Instance Level: Outcomes and Insights from the 2026 \textsc{Bright} Challenge}
%
%
%

\author{Hongruixuan Chen\dag, He Huang\dag, Haifeng Wang\dag, Jian Song, Junjue Wang, Weihao Xuan, Hamish Mitchell, Jiepan Li, Wei He, Liangpei Zhang, Zijie Wang, Chen Zhong, Jiazhen Zhao, Lei Hu, Ting Hu, Hongyan Zhang, Gregory Angelides, Miriam Cha, Clifford Broni-Bediako, Junshi Xia, Taylor Perron, Naoto Yokoya*

\thanks{\dag: H. Chen, H. Huang, and H. Wang contributed equally to this work.

Hongruixuan Chen, Jian Song, C. Broni-Bediako, Junshi Xia, and Naoto Yokoya are with the RIKEN Center for Advanced Intelligence Project (AIP), RIKEN, Tokyo 103-0027, Japan (e-mail: \nolinkurl{qschrx@gmail.com}, \nolinkurl{songjianrs@gmail.com}, \nolinkurl{akosabroni@gmail.com}, \nolinkurl{junshi.xia@riken.jp}, \nolinkurl{naoto.yokoya@riken.jp}).

Junjue Wang, Weihao Xuan, and Naoto Yokoya are with the Graduate School of Frontier Sciences, The University of Tokyo, Chiba 277-8561, Japan (e-mail: \nolinkurl{kingdrone@edu.k.u-tokyo.ac.jp}, \nolinkurl{weihaoxuan@g.ecc.u-tokyo.ac.jp}, \nolinkurl{yokoya@k.u-tokyo.ac.jp})

He Huang, Jiepan Li, Wei He, Liangpei Zhang, Haifeng Wang, Zijie Wang, Chen Zhong, Jiazhen Zhao, Lei Hu, and Hongyan Zhang are with the State Key Laboratory of Information Engineering in Surveying,
Mapping, and Remote Sensing, Wuhan University, Wuhan 430079, China (e-mail: \nolinkurl{huang_he@whu.edu.cn}, \nolinkurl{jiepanli@whu.edu.cn}, \nolinkurl{weihe1990@whu.edu.cn}, \nolinkurl{zlp62@whu.edu.cn}, \nolinkurl{wanghaifeng68@whu.edu.cn}, \nolinkurl{zijie.wang@whu.edu.cn},\nolinkurl{chen_zhong@whu.edu.cn}, \nolinkurl{zjz-whu@whu.edu.cn}, \nolinkurl{hulei.eva@whu.edu.cn}, \nolinkurl{zhanghongyan@cug.edu.cn}).

Ting Hu is with the School of Remote Sensing \& Geomatics Engineering, Nanjing University of Information Science and Technology, Nanjing 210044, China (e-mail: \nolinkurl{hutingrs@nuist.edu.cn}).

Hamish Mitchell and Taylor Perron are with the Department of Earth, Atmospheric and Planetary Sciences at MIT. Gregory Angelides and Miriam Cha are with MIT Lincoln Laboratory (email:\nolinkurl{whamitch@mit.edu}, \nolinkurl{perron@mit.edu}, \nolinkurl{gregangelides@ll.mit.edu}, \nolinkurl{miriam.cha@ll.mit.edu}).}
}

%
%

\markboth{Preprint. This work has been submitted to the IEEE for possible publication.}%
{Chen \MakeLowercase{\textit{et al.}}: Outcomes and Insights from the 2026 \textsc{Bright} Challenge}
%



\maketitle

\begin{abstract}
Rapid post-disaster response requires timely, building-level information on whether structures remain intact, are damaged, or are destroyed. Post-event optical imagery, however, may be unavailable because of cloud, smoke, or darkness. The \textsc{Bright} Challenge evaluated all-weather building damage mapping from a submeter-resolution pre-event optical image and a post-event SAR image. Participants were required to detect and delineate each building and assign exactly one of three mutually exclusive damage labels. The challenge extended the globally distributed \textsc{Bright} dataset with instance-level annotations for about 291,000 buildings across 16 disaster events spanning seven disaster types. The final phase was evaluated exclusively on two 2025 events absent from training: a wildfire event in California and a hurricane in Jamaica. A total of 157 participants made 1,289 submissions, and 46 teams entered the final phase. The two winning solutions achieved test mAPs of 0.182 and 0.181, approximately 8.7 times the public baseline of 0.021, but remained far below the best in-domain holdout score of 0.513. Across teams ranked in both phases, performance declined sharply and the rank order changed substantially. The two leading solutions independently favored modality-specific encoding, staged or late optical--SAR fusion, and an optical-dominant separation of building localization from damage recognition. The winning method additionally used scene-aware threshold adjustment and pseudo-label adaptation. These results identify cross-event generalization and stable severity discrimination as the principal remaining challenges. All data, annotations, baseline code, and winning solutions are publicly available at \url{https://github.com/ChenHongruixuan/BRIGHT}.
\end{abstract}


%
\IEEEpeerreviewmaketitle

\section{Introduction}

\IEEEPARstart{N}{atural} and human-made disasters damage buildings across the world every year. After an event, response teams need to know quickly which buildings are still usable, which are damaged, and which are destroyed. This information guides search and rescue, the allocation of resources, and recovery planning. Satellite imagery is the practical way to answer it over a whole affected region soon after an event, including places that are hard or unsafe to reach on the ground~\cite{voigt2016global}.

\par Most existing building damage mapping methods rely on optical imagery, often using paired pre-event and post-event optical images, and benchmarks evaluate damage at the pixel level~\cite{dong2013comprehensive, gupta2019xbd}. Two developments make a more operationally useful setting possible. First, very-high-resolution optical imagery can resolve individual buildings, enabling instance-level outputs in which each building is detected, outlined, and assigned a damage class. Such outputs are closer to response needs than dense damage rasters because they provide building counts, footprints, and geolocated objects that can be linked to downstream decision systems. Second, synthetic aperture radar (SAR) provides observations independent of cloud cover and daylight~\cite{plank2014rapid}. Pairing a pre-event optical image with a post-event SAR image therefore offers a realistic all-weather configuration for rapid response, although the cross-modal appearance gap makes the task substantially harder than optical-only comparison.

\begin{figure}[!t]
\centering
\includegraphics[width=0.485\textwidth]{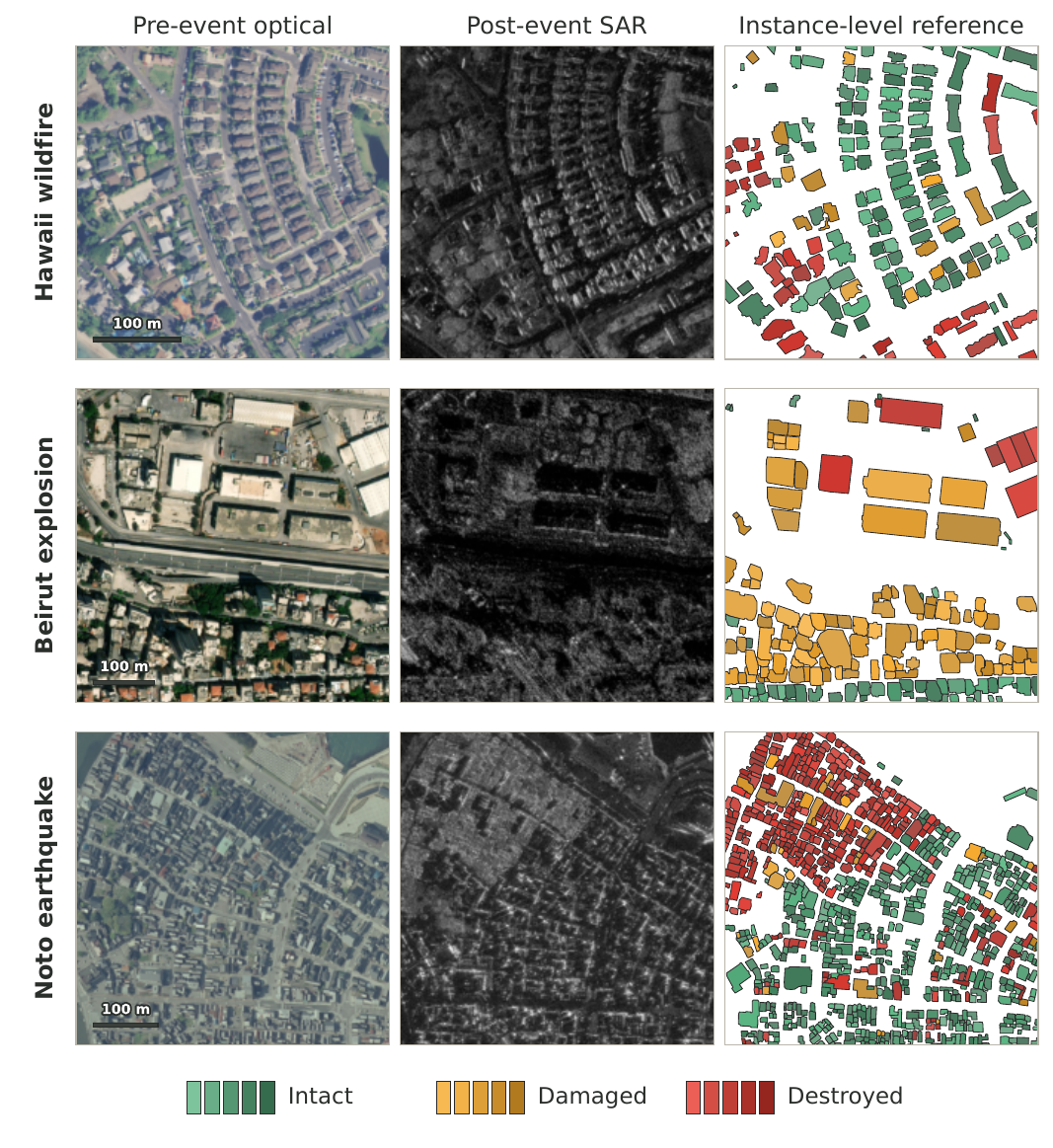}
\caption{Example annotated samples from the \textsc{Bright} training set, for three events. Each row shows the pre-event optical image, the post-event SAR image of the same scene, and the instance-level reference, in which every building is outlined and labeled as intact, damaged, or destroyed. Within a class, buildings are drawn in slightly different shades so that individual instances stay visible.}
\label{fig:BRIGHT_samples}
\end{figure}

\begin{table*}[!t]
\centering
\caption{Relevant community challenges and the positioning of the \textsc{Bright} Challenge. Year denotes the competition year. A held-out event/AOI test uses final-test events or areas that are
disjoint from those used for training.}
\label{tab:related_challenges}
\footnotesize
\setlength{\tabcolsep}{5pt}
\begin{tabular}{llllcc}
\toprule
Challenge & Year & Input imagery & Output granularity & Damage classes & Unseen-event test \\
\midrule
xView2~\cite{gupta2019xbd} & 2019 & Pre- \& post-event optical (VHR) & Pixel & 4 & $\times$ \\
ETCI Flood Detection~\cite{etci2021flood} & 2021 & Post-event SAR (Sentinel-1) & Pixel & -- & $\checkmark$ \\
SpaceNet-8~\cite{hansch2022spacenet8} & 2022 & Pre- \& post-event optical (VHR) & Pixel & 2 & $\times$ \\
Landslide4Sense~\cite{ghorbanzadeh2022landslide4sense, ghorbanzadeh2022l4soutcome} & 2022 & Post-event multispectral $+$ DEM/slope & Pixel & -- & $\checkmark$ \\
DFC 2023 Track 1~\cite{Liu2024Large} & 2023 & Optical $+$ SAR (VHR) & Instance  & -- & $\times$ \\
AI for Earthquake Response~\cite{ebel2026ai4earthquake} & 2025 & Pre- \& post-event optical (VHR) & Pixel & 2 & $\checkmark$ \\
DFC 2025 Track 2~\cite{persello2025dfc, liu2026dfcoutcome} & 2025 & Pre-event optical $+$ post-event SAR (VHR) & Pixel & 3 & $\checkmark$ \\
\textsc{Bright} Challenge & 2026 & Pre-event optical $+$ post-event SAR (VHR) & Instance & 3 & $\checkmark$  \\
\bottomrule
\end{tabular}
\end{table*}

\par A separate issue is how performance is evaluated. A model that performs well on held-out images from familiar disasters may not perform well on a new disaster, where the location, hazard type, season, sensor conditions, and damage patterns may differ from the training data. This distinction is central for deployment: operational use almost always concerns an event that the model has not seen~\cite{ebel2026ai4earthquake, liu2026dfcoutcome}. Evaluation protocols that do not separate in-domain performance from cross-event transfer can therefore overstate operational readiness.

\par The \textsc{Bright} Challenge was designed to evaluate this setting directly. As shown in Fig.~\ref{fig:BRIGHT_samples}, participants are given a pre-event optical image and a post-event SAR image, both with sub-meter-level spatial resolution, and need to produce an instance-level damage map in which every building is detected, delineated, and classified as intact, damaged, or destroyed. The evaluation separates two regimes. The development phase measures performance on a hidden holdout drawn from the training events, while the final test phase measures performance on events absent from the training data. This design allows the challenge to distinguish accuracy on familiar events from transfer to genuinely unseen events.

\par Several previous community benchmarks address related components of this problem. Table~\ref{tab:related_challenges} summarizes these community evaluations. The xView2 Challenge established large-scale building damage assessment from bi-temporal optical imagery as a semantic segmentation task~\cite{gupta2019xbd}. SpaceNet-6 pioneered building footprint extraction from combined SAR and optical imagery~\cite{shermeyer2020spacenet6}, and SpaceNet-8~\cite{hansch2022spacenet8} and SpaceNet-9~\cite{roony2026spacenet9} studied flood mapping and the co-registration challenges of multi-date, multi-sensor satellite imagery. Landslide4Sense and the ETCI flood detection competition provided community evaluations for specific hazard-mapping tasks~\cite{ghorbanzadeh2022landslide4sense, ghorbanzadeh2022l4soutcome, etci2021flood}. The 2023 IEEE GRSS Data Fusion Contest addressed instance-level building extraction and fine-grained roof classification from optical and SAR imagery, although without damage assessment~\cite{Liu2024Large}. The ESA and International Charter AI for Earthquake Response Challenge~\cite{ebel2026ai4earthquake} emphasized operational building damage assessment from very-high-resolution optical imagery. Closest to the \textsc{Bright} Challenge, the 2025 IEEE GRSS Data Fusion Contest~\cite{liu2026dfcoutcome, persello2025dfc, persello2025dfcreport} introduced paired optical and SAR imagery and out-of-domain evaluation for building damage mapping, but its task was formulated at the pixel level. The \textsc{Bright} Challenge combines their individual dimensions by evaluating all-weather, multimodal, cross-event building damage mapping at the instance level.

\par This paper reports the design and the outcome of the challenge in the spirit of an outcomes-and-insights report~\cite{ebel2026ai4earthquake, ghorbanzadeh2022l4soutcome, liu2026dfcoutcome}. Its contributions are threefold:

\begin{enumerate}
    \item \textbf{A new task with open resources.} We introduce the first community evaluation of all-weather building damage mapping formulated at the instance level, in which each building must be detected, delineated, and assigned a damage class from a pre-event optical and a post-event SAR image. We release the resources needed to work on this task: instance-level annotations that extend the \textsc{Bright} dataset, a public multimodal baseline with training code and weights, and a two-phase evaluation protocol that separates in-domain accuracy from cross-event transfer.
    \item \textbf{A summary of strong solutions.} We analyze the submissions of 46 teams and document the methods of the two winning teams. Despite being developed independently, these solutions converge on shared design decisions, namely modality-specific encoding with staged fusion, decoupling building localization from damage classification, and explicit adaptation to the target events.
    \item \textbf{Insights for future research.} We distill the lessons of the challenge for instance-level damage mapping, the clearest of which is that in-domain accuracy does not predict cross-event performance: the development-phase leaders were not the teams that won on the unseen events, and every team dropped sharply between the two phases. We discuss what this implies for method design, evaluation protocols, and benchmark construction.
\end{enumerate}

\par The remainder of this paper describes the data, task, and baseline (Section~\ref{sec:data_baseline}), the competition setup and results (Section~\ref{sec:competition_setup}), the two winning solutions (Sections~\ref{sec:first_solution} and~\ref{sec:second_solution}), and the lessons we draw from them (Section~\ref{sec:takeaways}).

\section{Data and Baseline}\label{sec:data_baseline}
\subsection{The \textsc{Bright} Dataset}

\begin{figure*}[!t]
\centering
\includegraphics[width=\textwidth]{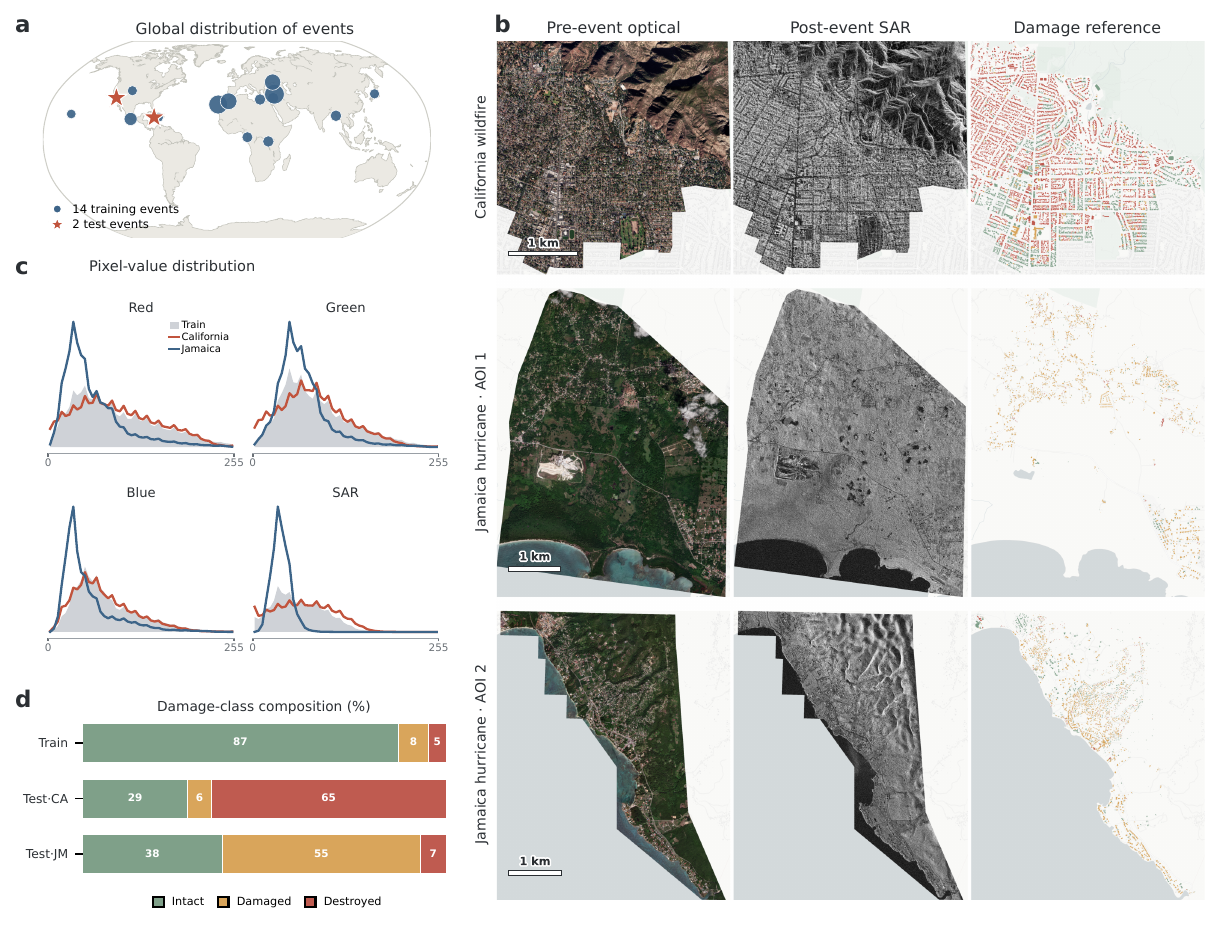}
\caption{Overview of the \textsc{Bright} challenge data. (a) Global distribution of the 14 training events (circles, sized by tile count) and the two unseen test events (red stars). (b) The three test areas, each shown as pre-event optical, post-event SAR, and the instance-level damage reference. (c) Pixel-value distributions of the optical and SAR channels for the training set and the two test events. (d) Damage-class composition of the training set and the two test events. }
\label{fig:BRIGHT_overview}
\end{figure*}

\par The challenge is built on the \textsc{Bright} dataset~\cite{chen2025bright}, a globally distributed, multimodal benchmark for building damage mapping. For each scene it pairs a pre-event optical image with a post-event SAR image, both at very high resolution, better than one meter per pixel. This resolution makes single buildings visible and instance-level labeling meaningful, and the post-event SAR input lets the task work through cloud and at night, where optical imagery alone would fail. The dataset covers 14 real disaster events across several continents and spans seven disaster types: earthquakes, floods, hurricanes, wildfires, volcanic eruptions, explosions, and armed conflict. Their global spread is shown in Fig.~\ref{fig:BRIGHT_overview}-(a). 

\par The original \textsc{Bright} dataset provides pixel-level damage labels. For this challenge we extended it to instance-level annotations: every building is labeled as a separate object, with its own footprint polygon and a single damage class. We use three classes that follow common practice: intact, damaged, and destroyed. The annotations are stored in COCO instance segmentation format~\cite{lin2014coco}, so that standard tools and metrics for instance segmentation from the field of computer vision can be used directly. Fig.~\ref{fig:BRIGHT_samples} shows example samples, with the paired optical and SAR inputs and the instance-level reference in which each building is outlined and classified. The \textsc{Bright} data and the new instance annotations are released for research use\footnote{\url{https://github.com/ChenHongruixuan/BRIGHT}}.

\subsection{Data Splits}
\begin{table}[t]
\centering
\caption{Data splits of the \textsc{Bright} challenge.}
\label{tab:splits}
\begin{tabular}{lrrr}
\toprule
Split & Tiles & Building instances & Labels \\
\midrule
Training and validation        & 3{,}029 & 244{,}976 & Released \\
Holdout (development phase)     & 366     & 32,840  & Hidden \\
Cross-event test (test phase)   & 426     & 13{,}384  & Hidden \\
\bottomrule
\end{tabular}
\end{table}
\par The data is divided into three parts, summarized in Table~\ref{tab:splits}. The training and validation set contains 3,029 image tiles and 244,976 labeled building instances, and its labels are released to participants. The holdout set contains 366 tiles and 32,840 building instances and is used in the development phase, whose labels were never released during the competition. The cross-event test set contains 426 tiles and 13,384 building instances and is used only in the final phase, again with hidden labels.

The cross-event test set is the most important part of the evaluation, because it more closely approximates a real deployment. It is made of two disasters that appear nowhere in the training, validation, or holdout data: a 2025 wildfire in California and a 2025 hurricane in Jamaica, marked by the red stars in Fig.~\ref{fig:BRIGHT_overview}-(a). Fig.~\ref{fig:BRIGHT_overview}-(b) shows the three test areas, each as a pre-event optical image, a post-event SAR image, and the instance-level damage reference. The two events have very different damage profiles, as the class composition in Fig.~\ref{fig:BRIGHT_overview}-(d) makes clear: in the wildfire scene most affected buildings are destroyed (4,729 of 7,321 instances), while in the hurricane scene most are damaged but still standing (3,307 of 6,063 instances). The two events also look different from the training data at the pixel level. Their optical and SAR value distributions depart from those of the training set (Fig.~\ref{fig:BRIGHT_overview}-(c)), so for a model the test data is new in both its damage pattern and its image statistics. This contrast lets us see whether a model trained on a mix of past events can adapt to the specific pattern of a new one.

\par The two events also differ in building density. The wildfire scene has 7,321 instances in 104 tiles, while the hurricane scene has 6,063 instances in 322 tiles, so the wildfire tiles are much denser.

\subsection{Evaluation Metric}
\par Submissions are scored with the standard COCO instance segmentation metrics~\cite{lin2014coco}. The main ranking metric is the mean average precision (mAP), averaged over intersection-over-union thresholds from 0.50 to 0.95 and over the three damage classes. We also report the average precision at the 0.50 and 0.75 thresholds (AP50 and AP75) and the average precision of each class. All metrics are computed on the server against the hidden labels.

\subsection{Baseline}
\par We provide a public baseline so that participants have a clear starting point and a common reference score. The baseline is a Mask R-CNN instance segmentation model~\cite{he2017mask} adapted to take two inputs, the pre-event optical image and the post-event SAR image. It is trained for 100 epochs with the default configuration that we release together with the training and inference code and the trained weights.

\par On the holdout set, the baseline reaches an mAP of 0.184, with an AP50 of 0.336 and an AP75 of 0.186. Its per-class average precision is 0.307 for intact, 0.103 for damaged, and 0.142 for destroyed, and its validation score is similar, at 0.185. On the cross-event test set, the same baseline reaches an mAP of 0.021, with per-class scores of 0.055, 0.004, and 0.003. The large drop from holdout to test shows that the task is hard to transfer even for a fixed model. It also sets a low reference that the teams had to beat on the unseen events.

\section{Submissions and Results}\label{sec:competition_setup}
\subsection{Competition Setup}

\begin{figure}[!t]
\centering
\includegraphics[width=0.485\textwidth]{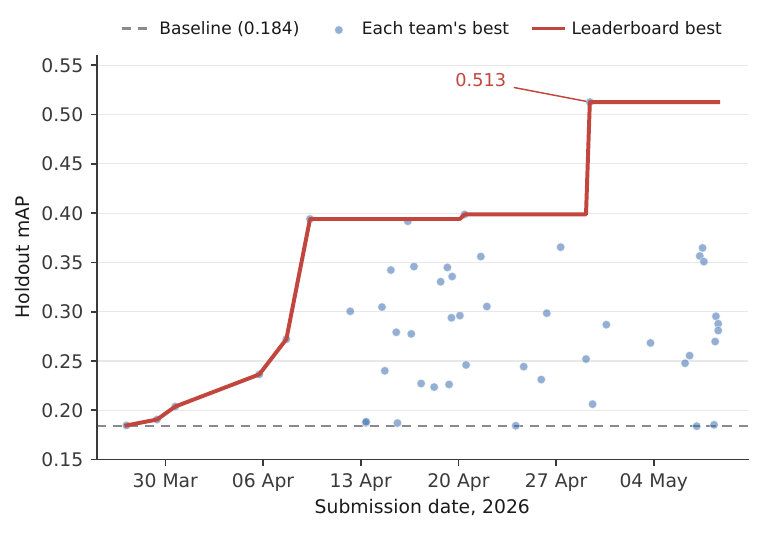}
\caption{Development phase. Each point is a team's best mAP on the hidden holdout split, plotted against submission date. The red line is the running leaderboard best.}
\label{fig:development_progress}
\end{figure}

\par The challenge was hosted on the CodaBench platform\footnote{\url{https://www.codabench.org/competitions/15134}}~\cite{xu2022codabench}, and participation was open to everyone with no registration fee. Teams downloaded the training and validation data, trained their models, and uploaded predictions as a COCO-format JSON file, with one record per detected building that includes a confidence score and a segmentation mask. The evaluation ran automatically on the server. Teams could use the \textsc{Bright} training data and any publicly available pre-trained model or auxiliary data, and any extra data had to be declared in the technical report.

\par In the development phase, each team could submit up to 20 times per day and was ranked on the holdout set. The test images were released without labels on 8 May 2026 for the test phase. In that phase, each team could submit up to 10 times in total, and the final ranking was based only on the test mAP. Teams that finished at the top had to share reproducible code and a short technical report. The final standings reflect the teams that met this requirement. The data was released on 25 March 2026, the final submission deadline was 15 May 2026, the results were announced on 27 May 2026, and the winners presented at the CVPR 2026 MONTI workshop\footnote{\url{https://sites.google.com/view/monti2026/home}} in June 2026.

\par In total, 157 participants took part and sent in 1,289 submissions. 60 teams appeared on the development leaderboard, and 46 teams remained active in the final test phase. Participants came from many countries and from both universities and industry.

\subsection{Development Phase Results}

\par As shown in Fig.~\ref{fig:development_progress}, in the development phase, most teams improved quickly over the baseline. The baseline mAP on the holdout set is 0.184. Early on, a large group of teams clustered just above this value, and the strongest teams then pulled far ahead. The best team reached an mAP of 0.513, about 2.8 times the baseline. As the next subsection shows, however, most of this in-domain gain did not carry over to the unseen test events.

\subsection{Test Phase Results}
\begin{figure}[!t]
\centering
\includegraphics[width=0.485\textwidth]{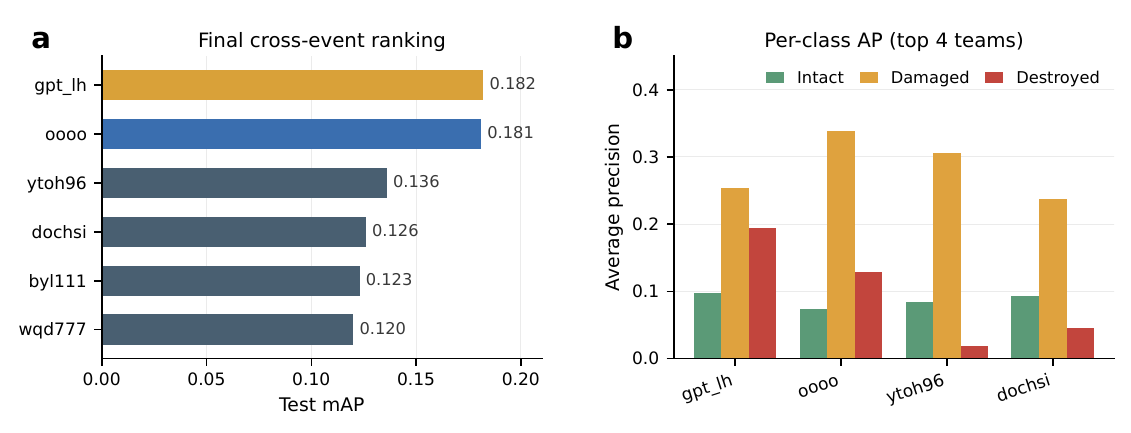}
\caption{Test phase, on the two unseen disasters. (a) Final ranking of the leading teams by test mAP. (b) Per-class average precision for the top four teams.}
\label{fig:test_accuracy}
\end{figure}

\begin{figure}[!ht]
\centering
\includegraphics[width=0.485\textwidth]{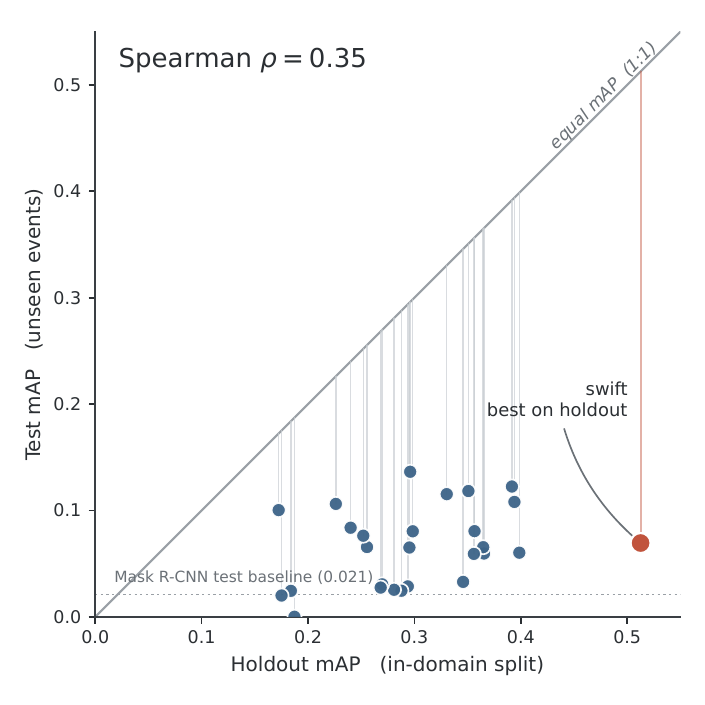}
\caption{In-domain versus cross-event accuracy. Each point is one of the teams that posted a ranked score in both phases, placed by its holdout mAP (in-domain split) and its test mAP (the two unseen events). The diagonal marks equal scores on the two sets. Every team falls below the diagonal, scoring lower on the unseen events than on the holdout split, and the rank order shifts between the phases (Spearman $\rho = 0.35$). The holdout winner (swift, highlighted) fell from 0.513 to 0.069. The dotted line is the Mask R-CNN test baseline (0.021).}
\label{fig:in_vs_cross}
\end{figure}

\begin{table}[!t]
\centering
\caption{The two winning teams of the \textsc{Bright} Challenge and their open-source solutions.}
\label{tab:winners}
\setlength{\tabcolsep}{4pt}
\begin{tabular}{@{}clp{2.8cm}p{3.8cm}@{}}
\toprule
Rank & Team & Members & Code \\
\midrule
1st & gpt\_lh & He Huang, Jiepan Li, Wei He, Liangpei Zhang & \url{https://github.com/huang-he99/BightSolution} \\
\addlinespace[2pt]
2nd & oooo & Haifeng Wang, Zijie Wang, Chen Zhong, Jiazhen Zhao, Lei Hu, Ting Hu, Hongyan Zhang & https://github.com/whf-68/Damage-Aware-SAR-Optical-Query-Learning-framework \\
\bottomrule
\end{tabular}
\end{table}
\par The test phase tells a different story. On the two unseen events, every team scored far below its holdout result. Fig.~\ref{fig:test_accuracy}-(a) gives the final ranking on the two unseen events, and Fig.~\ref{fig:test_accuracy}-(b) shows the top teams with their per-class scores. The two winning teams, $\text{gpt\_lh}$ and $\text{oooo}$, reached an mAP of 0.182 and 0.181. Table~\ref{tab:winners} lists their members and open-source solutions, and their methods are described in Sections~\ref{sec:first_solution} and~\ref{sec:second_solution}.

\par Three observations make the size of the generalization gap clear. First, the best holdout score ($\text{swift}$, 0.513) and the best test score ($\text{gpt\_lh}$, 0.182) come from different teams, so comparing the two best scores already shows a drop of about 65 percent. Second, and more telling, the same teams drop sharply between the two phases. Fig.~\ref{fig:in_vs_cross} plots holdout mAP against test mAP for the teams that competed in both. Every point sits well below the line of equal performance. The teams that led the holdout phase fell by 70 to 86 percent on the test set, and the holdout winner went from 0.513 to 0.069. Third, the rank order changes a great deal. The Spearman rank correlation between the holdout and test rankings is only 0.346. Taken together, these results show that high in-domain accuracy does not predict cross-event transfer, and that part of the holdout performance came from fitting the specific events seen during development.

\par The baseline shows the same pattern in an even stronger form. Its holdout mAP of 0.184 falls to 0.021 on the test set. The winning teams, at about 0.18, are well above this, so they did learn damage patterns that carry over to new events. Even so, the absolute scores stay low, which is a clear sign that the task is not yet solved.

\par The next two sections describe the methods of the first- and second-place teams. Their shared and diverging design decisions are analyzed in Section~\ref{sec:takeaways}.

\section{First-Place Team}\label{sec:first_solution}
 
\subsection{Motivation}
 
\begin{figure}[!t]
\centering
\includegraphics[width=0.485\textwidth]{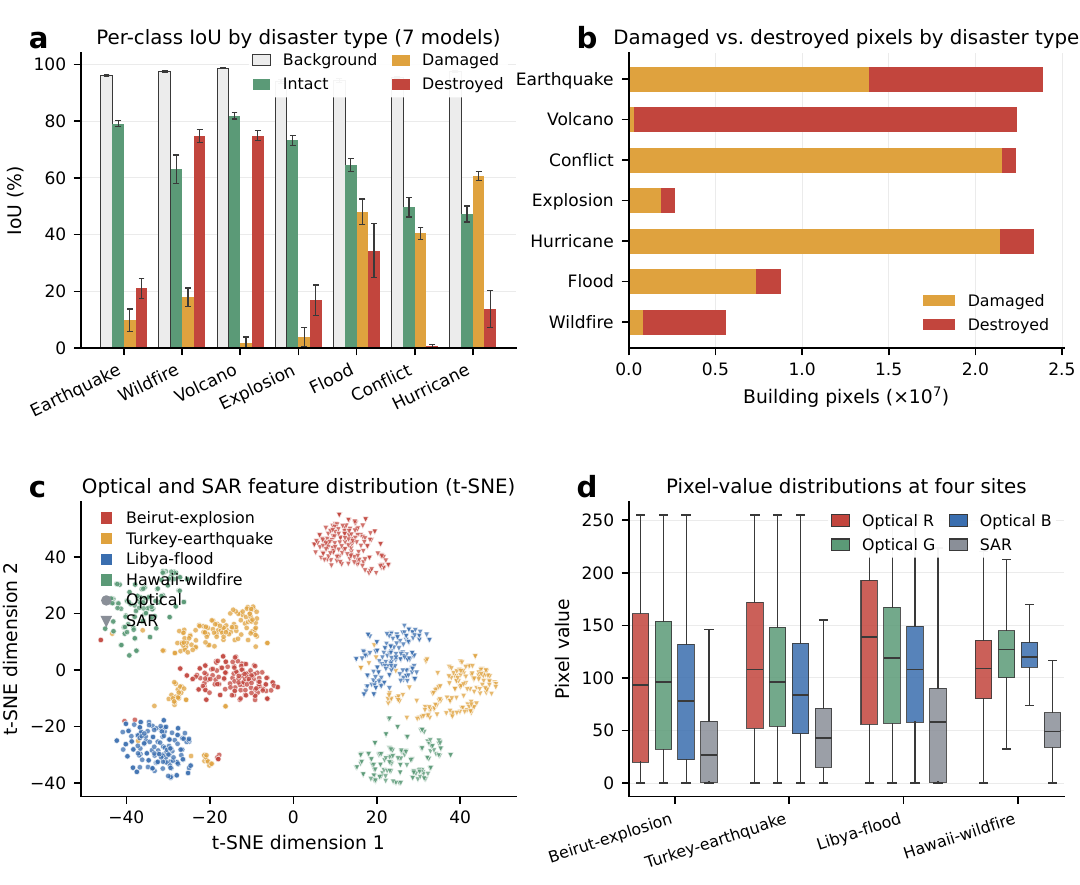}
\caption{Dataset heterogeneity and model performance variation across disaster types. (a) Per-class IoU of seven benchmark models across the seven disaster types of the \textsc{Bright} dataset; bars show the mean over the models and error bars one standard deviation. (b) Damaged and destroyed building pixels per disaster type, computed over the full training set. (c) t-SNE embedding of deep features of optical and SAR patches from four representative sites; color denotes the site and the marker denotes the modality. (d) Distributions of optical (R, G, B) and SAR pixel values at the same four sites; boxes show the median and interquartile range, and whiskers extend to 1.5 times the interquartile range.}
\label{fig:disaster_analysis}
\end{figure}
 
\par As shown by the IoU distributions of deep learning models across the seven disaster types of the \textsc{Bright} dataset~\cite{chen2025bright} (Fig.~\ref{fig:disaster_analysis}-(a)), detection performance exhibits marked unevenness in generalization capability and robustness across disaster scenarios, indicating distinct domain discrepancies. The first-place team attributed this variability to two primary factors:
\begin{itemize}
    \item \textbf{Imaging source and regional discrepancies:} As illustrated in Fig.~\ref{fig:disaster_analysis}-(c) and Fig.~\ref{fig:disaster_analysis}-(d), optical-SAR data pairs across different regions in the \textsc{Bright} dataset originate from diverse acquisition sources. Consequently, their feature representations are significantly confounded by regional factors, which include local geography and environmental conditions.
    \item \textbf{Disaster-specific damage patterns:} Different disaster types induce entirely distinct morphological characteristics and spatial distribution modes of building damage. For instance, considering the quantity of damaged buildings (Fig.~\ref{fig:disaster_analysis}-(b)), three separate disaster-response modes can be identified, namely Destroyed-dominant, Balanced, and Damage-dominant scenarios.
\end{itemize}
 
\par Indiscriminately lumping these heterogeneous samples together for joint training hinders the model from learning universally robust, generalized features. This naive approach triggers feature confusion and gradient interference among multiple tasks, thereby significantly suppressing the model's recognition accuracy on minority or low-signature disaster types. 
 
\par To address these challenges, the team proposed a framework named Scene-Segregated Pseudo-Label Learning for cross-modal instance-level building damage mapping. To adapt to diverse damage distributions, a Disaster Scene Classification module segregates the different scenarios. Based on the classification results, a Scene-Segregated Training Strategy trains a dual-stream two-stage instance-level change detection model, achieving fine-grained feature decoupling and high-accuracy adaptive mapping. Furthermore, to mitigate the adverse effects of imaging sources and regional variations, a Pseudo-Label Learning strategy leverages high-confidence pseudo-labels to guide the spatial alignment and self-supervised consistency learning of cross-modal and cross-regional features, effectively bridging the domain gap between different data sources. Details are provided in Section~\ref{sec:solution}.

\subsection{Solution}
\label{sec:solution}
 
\begin{figure}[!t]
\centering
\includegraphics[width=0.485\textwidth]{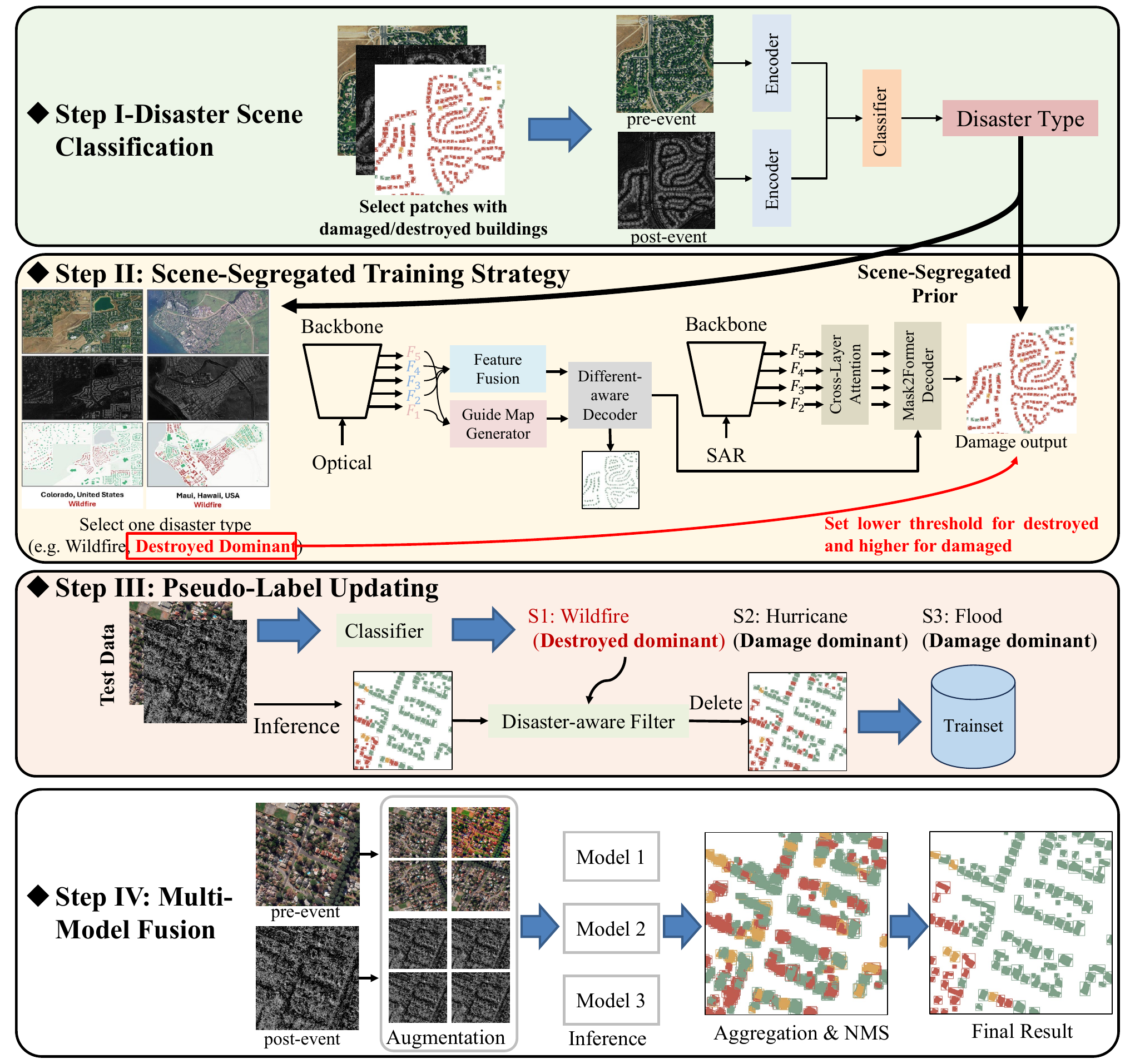}
\caption{Workflow of the first-place team's Scene-Segregated Pseudo-Label Learning framework for cross-modal instance-level building damage mapping. The pipeline graphically illustrates the sequential execution from (Step I) Disaster Scene Classification for scene prior generation and (Step II) Scene-Segregated Training Strategy utilizing adaptive threshold re-calibration to (Step III) Disaster-Aware Pseudo-Label Updating for domain gap mitigation and (Step IV) Test-Time Multi-Model Fusion governed by a customized Mask Non-Maximum Suppression (Mask NMS) algorithm.}
\label{fig:gpt_lh_framework}
\end{figure}

To systematically address the challenges of inter-scene damage heterogeneity and cross-modal domain gaps, the first-place team introduced the Scene-Segregated Pseudo-Label Learning framework~\cite{li2025building} for cross-modal instance-level building damage mapping\footnote{Code is available at \url{https://github.com/huang-he99/BightSolution.git}}. As illustrated in Fig.~\ref{fig:gpt_lh_framework}, the framework consists of four sequential stages: disaster scene classification, scene-segregated model training, pseudo-label updating, and test-time multi-model fusion.
 
\subsubsection{Disaster Scene Classification}
Disaster Scene Classification serves as a foundational step to handle macro-level scene variations before performing instance-level predictions \cite{zheng2020foreground}. Given a pair of co-registered patches consisting of pre-event optical imagery and post-event SAR data, the framework first selects regions containing damaged or destroyed buildings. These cross-modal image patches are concurrently fed into a dual-encoder network to extract complementary spatial and structural representations. The extracted features are aggregated and processed by a scene classifier to determine the specific disaster type. The classification output functions as a high-level scene prior, providing an informative guide map that allows the subsequent components to effectively separate and process different categories of disaster data.
 
 
\subsubsection{Scene-Segregated Training Strategy}
\par This strategy utilizes a dual-stream two-stage instance-level change detection network to dynamically adjust internal decision boundaries based on the identified disaster category. In the first stage, multi-level feature maps ($\mathrm{F}_{1}$ to $\mathrm{F}_{5}$) extracted from the pre-event optical backbone pass through a Feature Fusion module and a Guide Map Generator \cite{li2025towards, li2026progressive}. These features are then combined with the post-event building features via a Different-aware Decoder to produce a binary building footprint localization mask, which is denoted as the building output \cite{li2025overcoming}. 
 
\par In the second stage, the localized building footprints and the fused optical features are correlated with the post-event SAR backbone features through a Cross-Layer Attention mechanism. The attended features are subsequently channeled into a Mask2Former \cite{cheng2022masked} Decoder to generate the final instance-level damage output. Crucially, to accommodate the skewed damage distributions inherent to different disaster types, the framework dynamically re-calibrates the classification thresholds within the decoder. For instance, when a scene is flagged as a wildfire, which typically represents a Destroyed-dominant scenario, the network adaptively reduces the decision threshold for the destroyed category while increasing the threshold for the damaged category, improving the model's sensitivity to severe structural destruction.

\subsubsection{Pseudo-Label Updating}
 
\par To mitigate performance degradation caused by domain shifts, the third stage incorporates an iterative Pseudo-Label Updating mechanism. During inference on unlabeled target domain test data, the trained dual-stream network generates initial instance-level predictions that serve as raw pseudo-labels. Concurrently, the test patches are processed by the scene classifier to determine their respective environmental contexts. To eliminate noise and false positives in the raw predictions, a Disaster-aware Filter is introduced. This filter evaluates the consistency between the predicted building damage states and the structural characteristics dictated by the macro disaster type. The remaining high-confidence, clean pseudo-labels are integrated back into the training dataset, allowing for self-supervised model retraining and continuous optimization of the feature space.
 
 
\subsubsection{Test-Time Multi-Model Fusion}
\par The final stage employs a Multi-Model Fusion strategy to improve robustness and stability during deployment. To capture diverse feature abstractions and prevent over-fitting, the team trained an ensemble of three distinct model variations. During test-time inference, the target cross-modal inputs are transformed into multiple augmented views, which are then processed in parallel across the three trained models to yield a collection of raw instance-level detections. To resolve spatial conflicts and eliminate redundant bounding boxes, an aggregation and mask processing stage is governed by a customized Mask Non-Maximum Suppression (Mask NMS) algorithm \cite{wang2020solov2}. The Mask NMS suppresses overlapping instance boundaries, reconciles conflicting damage category assignments, and fuses the complementary advantages of each individual model, thereby delivering the final high-precision instance-level building damage map.
 
 

\subsection{Results}
\label{sec:experiments}
 
This subsection presents the experimental evaluation of the Scene-Segregated Pseudo-Label Learning framework, covering the implementation setup, the architectures used for evaluation, and the quantitative and qualitative results.
 
\subsubsection{Experimental Settings and Infrastructure}
To ensure reproducibility, the framework was implemented and evaluated under a standardized hardware and hyperparameter configuration. All models were trained and evaluated on a single NVIDIA GeForce RTX 4090 GPU (24GB) for 100 epochs using the AdamW optimizer with an initial learning rate of $10^{-4}$. The batch size was set to 4 for the building footprint localization branch and 2 for the change detection branch, with input image patches resized to $1024 \times 1024$ pixels. To evaluate framework adaptability, Mask2Former \cite{cheng2022masked}, YOLOv10 \cite{wang2024yolov10}, and YOLO26 \cite{chakrabarty2026yolo26} were incorporated as the baseline detection decoders.
 
 
 
\subsubsection{Quantitative Comparison}
\par An ablation study was conducted to evaluate the contribution of each component within the framework, with performance measured by mAP on the cross-event test set. The results are compiled in Table~\ref{tab:quantitative_comparison}.

\begin{table}[!t]
\centering
\caption{Ablation of the first-place solution on the cross-event test set.}
\label{tab:quantitative_comparison}
\begin{tabular}{lccc}
\toprule
\textbf{Method Configuration} & \textbf{Pseudo-Label} & \textbf{Post-Process} & mAP \\ \midrule
Baseline   & $\times$              & Native                & 0.0409                 \\
+ Early Fusion                & $\times$              & Native                & 0.0542                 \\
+ Dual-Stage Fusion           & $\times$              & Native                & 0.1095                 \\
\textbf{Complete Solution} & $\checkmark$        & \textbf{Mask NMS}     & \textbf{0.1815}        \\ \bottomrule
\end{tabular}
\end{table}

\par As indicated in Table~\ref{tab:quantitative_comparison},
the vanilla baseline using only optical imagery yields a suboptimal mAP of 0.0409. Early fusion of cross-modal information marginally improves the metric to 0.0542, whereas the dual-stage fusion mechanism doubles the accuracy to 0.1095, demonstrating the clear advantage of decoupling structural localization from damage assessment. Incorporating iterative pseudo-label updates and Mask NMS post-processing allows the complete solution to achieve the highest mAP of 0.1815, which significantly outperforms the initial baseline and reinforces the efficacy of the joint scene-segregation and domain-bridging design.

\begin{figure*}[!t]
    \centering
\includegraphics[width=0.98\textwidth]{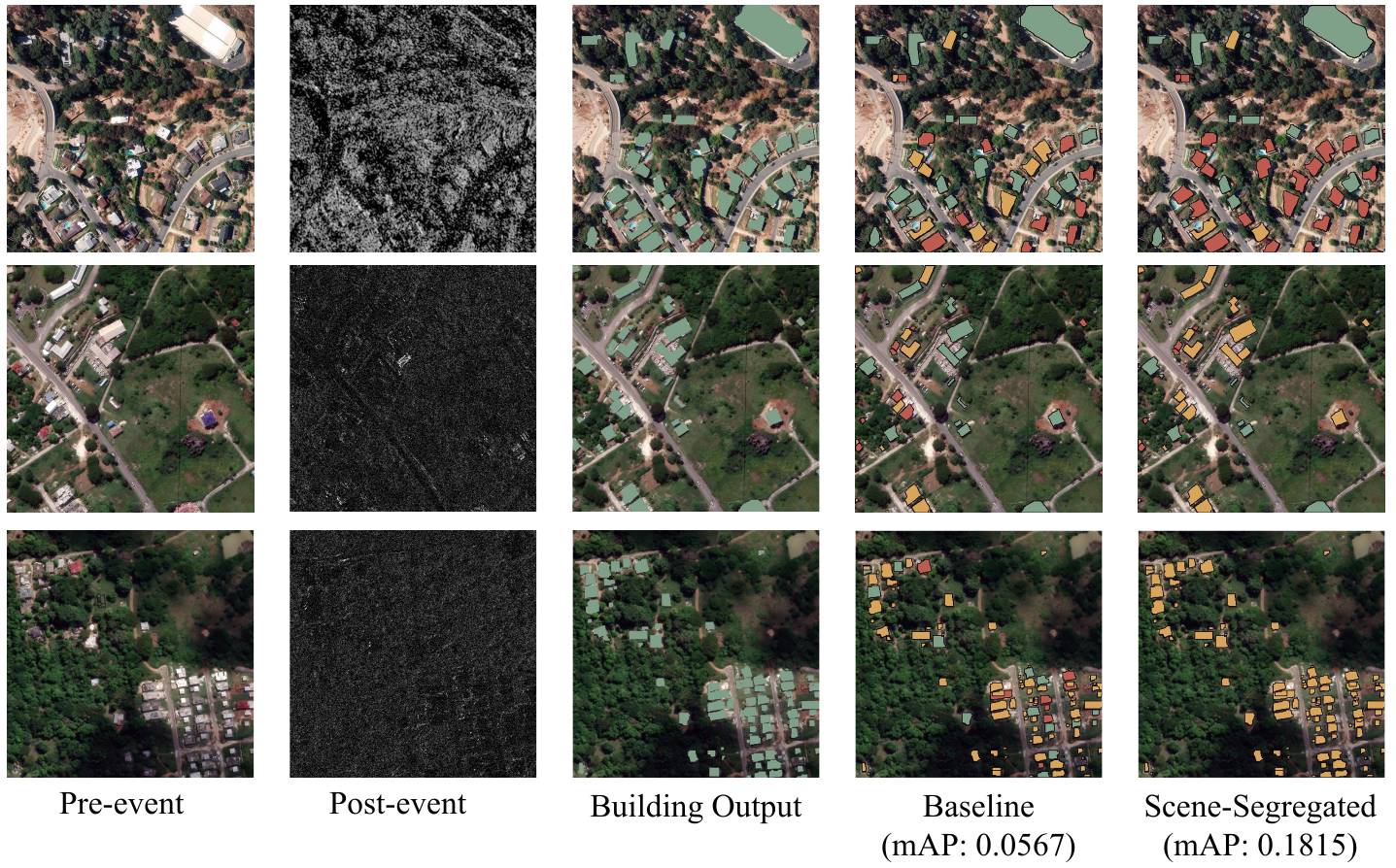} 
    \caption{Qualitative comparison of the first-place team's building damage mapping results. The columns from left to right represent the pre-event imagery, post-event imagery, intermediate building outputs, mixed-baseline predictions, and scene-segregated framework predictions.}
\label{fig:visual_results}
\end{figure*}
 
\subsubsection{Qualitative Analysis and Visual Results}
\par Fig.~\ref{fig:visual_results} presents a visual comparison of instance-level building damage mapping across diverse disaster scenes. From left to right, the columns sequentially display pre-event optical imagery, post-event SAR imagery, intermediate building footprint outputs, mixed-data baseline predictions (mAP 0.0567), and the scene-segregated results (mAP 0.1815). The conventional mixed baseline suffers from severe missing detections and misclassifications due to feature confusion across conflicting damage patterns. Conversely, the scene-segregated framework eliminates inter-class interference by leveraging scene-specific classification priors and adaptive thresholds, capturing precise boundaries for both damaged and destroyed instances.
 
\section{Second-Place Team}\label{sec:second_solution}
\subsection{Motivation}
\par As shown by the large gap between in-domain holdout performance and cross-event test performance, instance-level building damage mapping from paired optical and SAR imagery exhibits strong cross-event domain shift and severe class-wise imbalance. The second-place team therefore focused on making post-event SAR evidence useful for instance-level damage recognition while preserving the stable localization ability of an optical-dominant detector.
 
\par The team attributed this difficulty to two primary factors:
\begin{itemize}
    \item \textbf{The physical discrepancy between pre-event optical imagery and post-event SAR imagery.} Optical images preserve building footprints, texture, and neighborhood context before the disaster, whereas SAR images capture post-event scattering responses under smoke, cloud cover, and poor illumination. Directly concatenating them as four input channels forces early convolutional filters to mix optical appearance and SAR backscatter in one representation, which can weaken both modalities.
    \item \textbf{The ambiguity and rarity of the damaged category.} Damaged buildings are visually and semantically intermediate. They are often closer to intact buildings in pre-event optical appearance, while their SAR responses may overlap with both intact and destroyed states. This makes damaged recognition much less stable than intact or destroyed recognition, especially when the disaster type, imaging source, and geographic region change simultaneously.
\end{itemize}
 
\begin{figure*}[!t]
\centering
\includegraphics[width=0.98\textwidth]{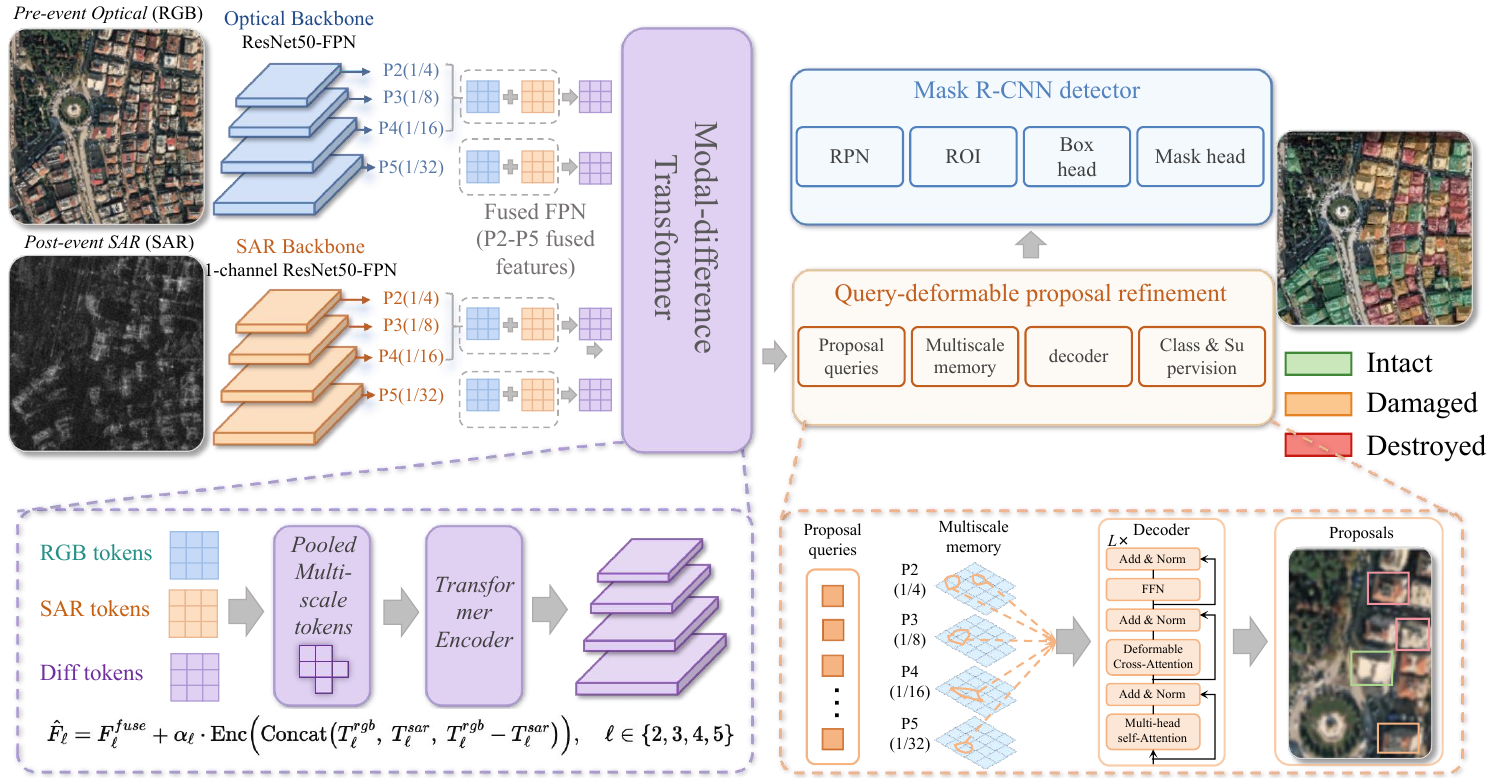}
\caption{Overview of the second-place team's Damage-Aware SAR-Optical Query Learning framework. Pre-event optical imagery and post-event SAR imagery are encoded by separate feature streams, fused across the feature pyramid, refined through cross-modal interaction and proposal-level query reasoning, and decoded by the instance segmentation heads.}
\label{fig:second_bright_model}
\end{figure*}
 
\par If the two modalities are fused too early and all damage states are optimized only through a standard instance segmentation objective, the model is prone to modality confusion, damaged-class under-calibration, and weak cross-event transfer. The solution was therefore designed to preserve modality-specific feature extraction while injecting explicit cross-modal difference cues and proposal-level damage reasoning.
 
\subsection{Solution}
\par The second-place solution can be summarized as \textit{Damage-Aware SAR-Optical Query Learning}. As shown in Fig.~\ref{fig:second_bright_model}, it uses a Mask R-CNN-style instance segmentation framework~\cite{he2017mask} as the detection backbone, but modifies the multimodal representation and proposal refinement stages. The input consists of paired pre-event RGB and post-event SAR imagery, but instead of sending the resulting four-channel tensor through one backbone, the model uses two ResNet-50-FPN streams~\cite{he2016resnet, lin2017fpn}: one for optical RGB and one adapted to single-channel SAR.
 
\par At each FPN level, optical and SAR features are concatenated and compressed by a $1 \times 1$ fusion layer. This layer is initialized as an optical passthrough, so training starts from a stable optical representation and gradually learns how much SAR information should be injected. This design is useful for noisy SAR scenes because it avoids forcing unstable SAR signatures into the detector too early.
 
\par The second component is a residual cross-modal difference interaction block. For selected pyramid levels, the model constructs compact tokens from optical features, SAR features, signed optical-SAR differences, and absolute differences. These tokens are encoded jointly and returned to the feature pyramid as a residual update:
\begin{equation}
    \mathrm{F}_l = \mathrm{F}_l^{\text{fuse}} + \alpha_l \cdot \mathrm{T}_l(\mathrm{F}_l^{\text{rgb}}, \mathrm{F}_l^{\text{sar}}),
\end{equation}
where $\alpha_l$ is a learnable residual scale initialized to a small value. This gives the detector a conservative path: when SAR evidence is unreliable, predictions can remain close to the optical-dominant feature; when SAR evidence is informative, the residual branch can emphasize cross-modal damage cues.
 
The third component is damage-aware query proposal refinement. Region proposals from the region proposal network (RPN)~\cite{ren2015faster} are converted into proposal queries containing proposal geometry, sampled multiscale features, and level embeddings. These queries attend to a compact multimodal memory and produce auxiliary class and mask predictions. During inference, the module refines proposal scores, labels, and masks in a residual manner. This keeps the localization stability of a region-based detector while allowing damage decisions to be made at the building-instance level, where weak SAR evidence can be aggregated more effectively.
 
The training protocol also included auxiliary SAR-optical representation learning and several implementation optimizations for dense scenes, including cached four-channel images, cached instance masks, crop-before-mask-loading, chunked RPN anchor matching, and memory-efficient mask loss computation.
 
\subsection{Results}
\begin{figure}[!t]
\centering
\includegraphics[width=0.485\textwidth]{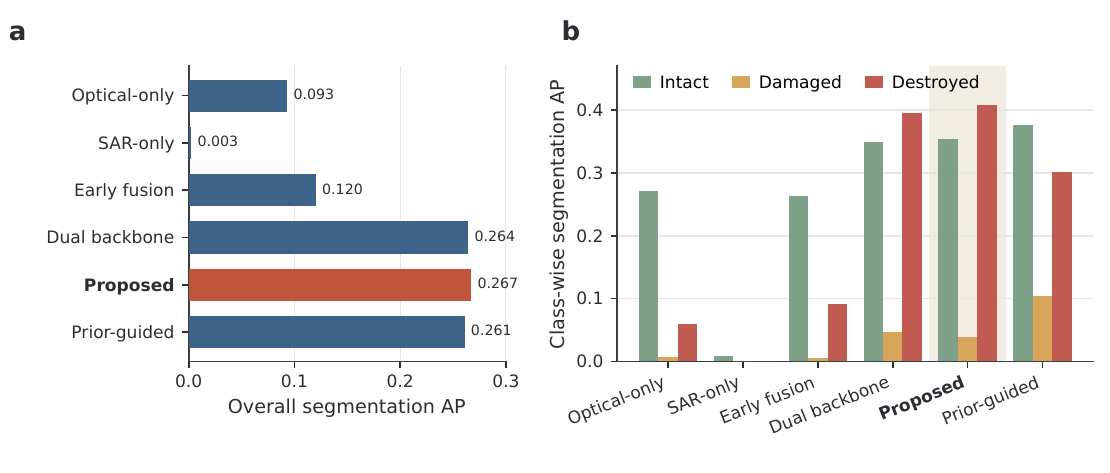}
\caption{Second-place team's model and class-wise AP synthesis. Overall AP improves strongly from single-modality or early-fusion settings to late dual-backbone fusion, while damaged AP remains much lower than intact and destroyed AP.}
\label{fig:second_model_class_ap}
\end{figure}

Under the team's standardized validation protocol, the proposed query-deformable dual-backbone model reached an mAP of 0.267, with an AP50 of 0.464 and an AP75 of 0.283. The class-wise AP was 0.354 for intact buildings, 0.038 for damaged buildings, and 0.409 for destroyed buildings. This result was clearly stronger than single-modality and early-fusion baselines: a SAR-only Mask R-CNN reached an mAP of only 0.002, an optical-only Mask R-CNN reached 0.093, and early RGB+SAR fusion reached 0.120. A late-fusion dual-backbone Mask R-CNN reached 0.264, showing that preserving modality-specific streams was the largest contributor, while query-level refinement provided an additional gain and the strongest destroyed-class AP. Fig.~\ref{fig:second_model_class_ap} visualizes this overall gain together with the remaining class-wise imbalance.
 
\begin{table}[!t]
\centering
\caption{Second-place team's main validation results under its standardized protocol.}
\label{tab:second_team_main}
\begin{tabular}{lcccc}
\toprule
Model & mAP & AP50 & AP75 & $\text{AP}_{\text{dmg}}$ \\
\midrule
Optical-only Mask R-CNN & 0.093 & -- & -- & 0.007 \\
Early RGB+SAR fusion & 0.120 & -- & -- & 0.006 \\
Late-fusion dual-backbone & 0.264 & -- & -- & 0.047 \\
Proposed query-deformable & 0.267 & 0.464 & 0.283 & 0.038 \\
\bottomrule
\end{tabular}
\end{table}
 
\par
\begin{figure}[!t]
\centering
\includegraphics[width=0.485\textwidth]{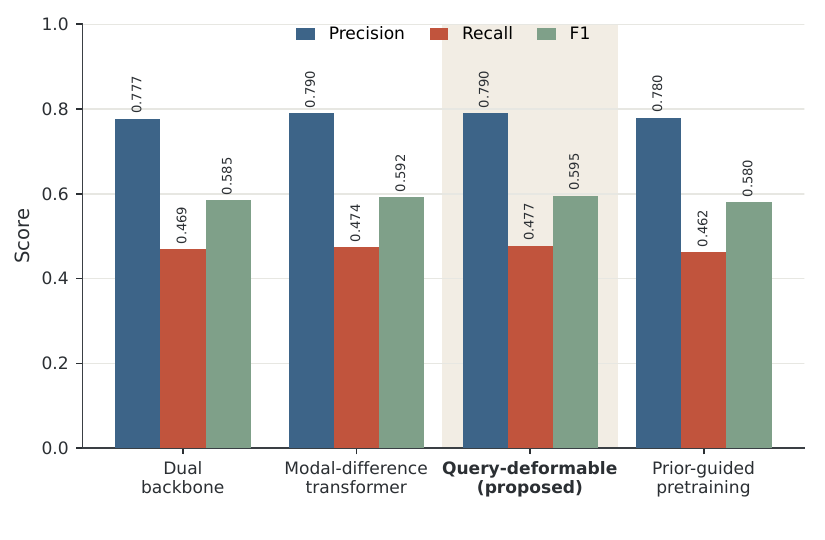}
\caption{Unified full-validation instance-matching summary for the second-place team's main variants. The query-deformable model gives the highest F1, but its margin over the modal-difference transformer is small, indicating that the leading multimodal variants form a close top tier.}
\label{fig:second_unified_matching}
\end{figure}

 The ablation study showed that several multimodal variants formed a close top tier. The proposed query-deformable model achieved an mAP of 0.267, followed by a modal-difference transformer variant at 0.265, the late-fusion dual-backbone at 0.264, and a prior-guided pretraining variant at 0.261. The prior-guided model achieved the best damaged AP, 0.105, but its destroyed AP dropped to 0.302, compared with 0.409 for the proposed query-deformable model. This indicates that improving the damaged class can shift the damaged--destroyed decision boundary rather than uniformly improving all classes. The unified instance-matching summary in Fig.~\ref{fig:second_unified_matching} supports the same reading: the query-deformable model gives the best F1 among the evaluated variants, but its margin over the modal-difference transformer is small.
 
\par
\begin{figure}[!t]
\centering
\includegraphics[width=0.485\textwidth]{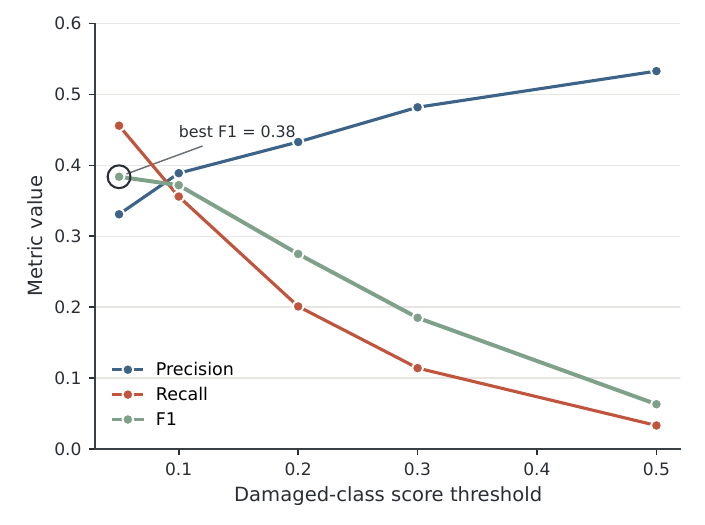}
\caption{Damaged-class precision--recall behavior. The steep tradeoff indicates that damaged recognition is governed by class-boundary calibration and rare-class recall, rather than by a single confidence threshold.}
\label{fig:second_damaged_pr}
\end{figure}

 The team also evaluated generalization under event-level domain shift. On an event-holdout validation split, the proposed query-deformable model improved mAP from 0.049 for the standard Mask R-CNN baseline and 0.064 for a damaged hard-negative-mining variant to 0.148. However, the damaged class remained near zero in this setting, confirming that damaged recognition was the least transferable part of the task. The damaged-class precision--recall behavior in Fig.~\ref{fig:second_damaged_pr} further shows that the bottleneck is a calibration problem rather than a single-threshold issue. In the final test phase of the challenge, the team ranked second with a test mAP of 0.181, only 0.001 below the first-place score.
 
\par
\begin{figure}[!t]
\centering
\includegraphics[width=0.485\textwidth]{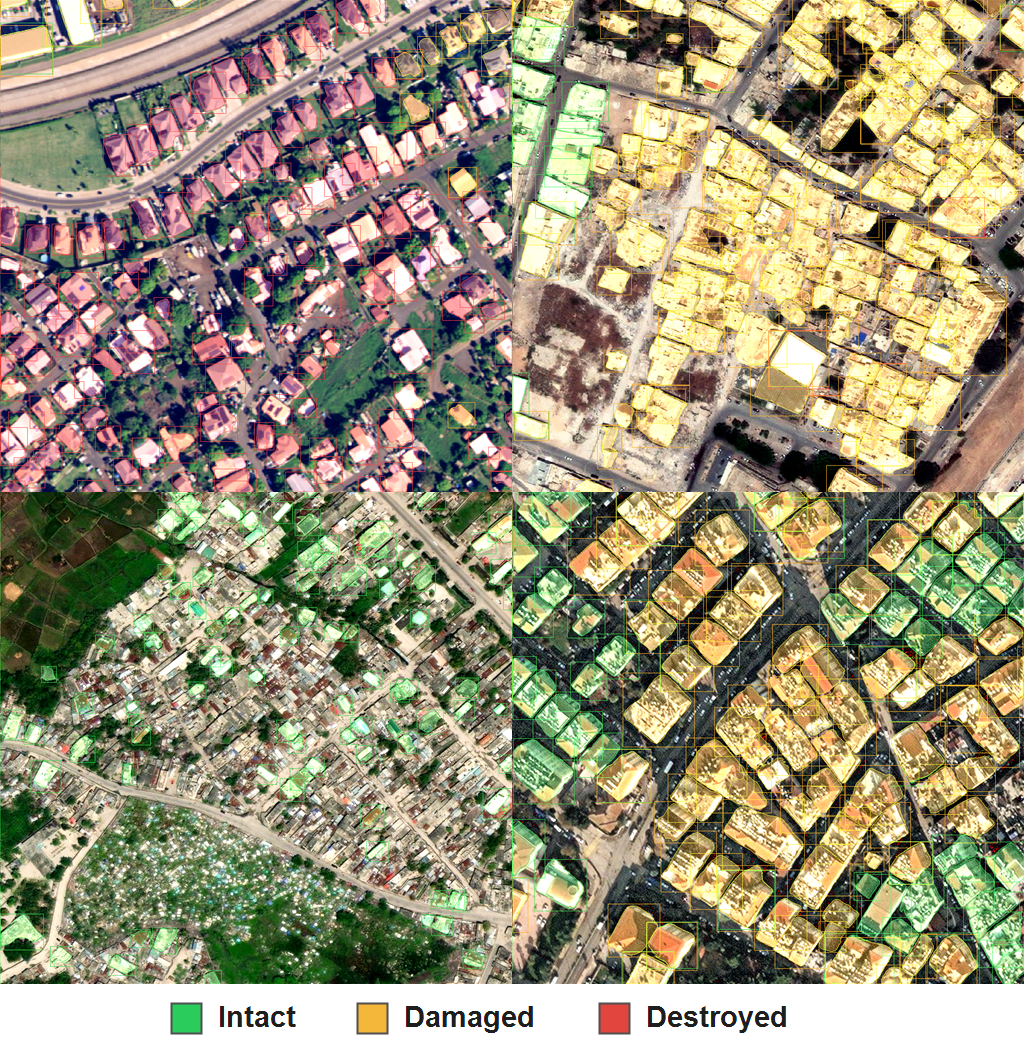}
\caption{Representative detection results of the second-place model across different disaster scenes. The legend indicates intact, damaged, and destroyed instances.}
\label{fig:second_detection_visualization}
\end{figure}

 Fig.~\ref{fig:second_detection_visualization} gives several representative detection visualizations. The examples show that the model can recover many building instances across different disaster scenes, while still missing some objects and shifting part of the class distribution toward damaged predictions.
 
\section{Discussions and Takeaways}
\label{sec:takeaways}
 
\par The challenge produced two complementary sources of evidence: blind test results from 46 teams under identical conditions, and the detailed reports of the two winning teams. This section combines both and distills what they imply for research on instance-level building damage mapping.
 
\subsection{Convergent Design Lessons from the Winning Solutions}
\par Although developed independently, the two winning solutions converge on two design decisions. The first is that optical-SAR fusion should not be treated as channel stacking. Both teams preserved modality-specific feature extraction and fused afterwards: the first-place team through a dual-stream two-stage network, and the second-place team through two backbones combined by a fusion layer initialized as an optical passthrough plus a residual cross-modal interaction. The ablations quantify the choice. For the first-place team, early fusion reached an mAP of 0.0542 on the cross-event test set against 0.1095 for the dual-stage design; for the second-place team, early fusion reached 0.120 under its validation protocol against 0.264 for the late-fusion dual-backbone, the single largest gain in its ablation.
 
\par The second shared decision is to decouple building localization from damage classification. In both systems, geometry is anchored in the pre-event optical image, where buildings are intact and clearly visible, and the post-event SAR image serves as evidence for the damage state rather than as an equal partner in detection: the first-place network localizes footprints in its first stage before assigning damage in the second, and the second-place detector starts from an optical-dominant representation and makes damage decisions through proposal-level query refinement. For researchers entering this task, this is the most transferable recipe of the challenge: obtain reliable footprints first, then treat severity assignment as a separate, explicitly calibrated decision.
 
\begin{figure*}[!t]
\centering
\includegraphics[width=0.935\textwidth]{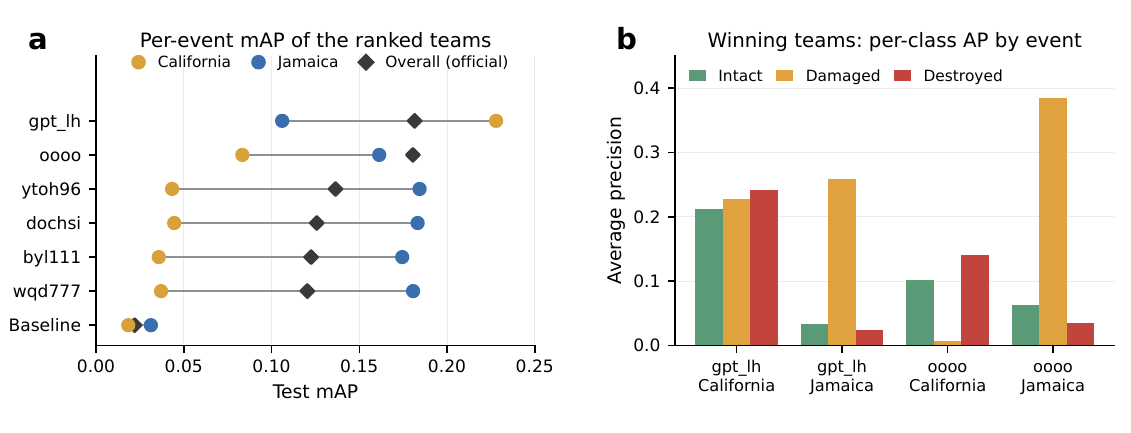}
\caption{Per-event breakdown of the final test scores. (a) Per-event mAP of the ranked teams and the public baseline; diamonds mark the official score on both events combined, which is a pooled score rather than the average of the two per-event values. (b) Per-class AP of the two winning teams on each event. }
\label{fig:per_test_event}
\end{figure*}

\subsection{In-Domain Accuracy Does Not Predict Cross-Event Transfer}
\par The clearest single outcome of the challenge is the generalization gap in Fig.~\ref{fig:in_vs_cross}. Every team scored far lower on the two unseen events than on the holdout split, the development-phase leaders dropped by 70 to 86 percent, the holdout winner fell from 0.513 to 0.069, and the rank correlation between the two phases was only 0.35. A substantial part of in-domain performance therefore reflects fitting the specific events available during development rather than learning transferable damage representations; with up to 20 submissions per day against a fixed holdout, some degree of overfitting to the leaderboard is unavoidable. The lesson for method development is to validate under event-holdout protocols, training on all events but one and validating on the one held out, in line with model-selection practice in domain generalization~\cite{gulrajani2021search}. The second-place team did exactly this and saw its validation mAP fall from 0.2670 under a standard split to 0.1482 under an event holdout, internally reproducing the phase gap that surprised many teams. The lesson for benchmark builders is symmetric: hidden, genuinely unseen events are not an optional refinement, because evaluation on familiar events systematically overstates operational readiness~\cite{chen2026earthobservationdisastermapping}.
 
\subsection{Severity Classes Are Unstable Across Events}
\par In-domain evidence and cross-event evidence tell opposite stories about the three severity classes. In the teams' own development experiments, damaged was consistently the weakest class. The second-place team measured a damaged AP of 0.038 against 0.354 for intact and 0.409 for destroyed under its validation protocol, saw damaged recognition fall to nearly zero under its event-holdout split, and found a precision--recall tradeoff so steep that no single confidence threshold resolves it (Fig.~\ref{fig:second_damaged_pr}); its prior-guided variant raised damaged AP to 0.105 only while dropping destroyed AP from 0.409 to 0.302, shifting the damaged--destroyed boundary rather than improving it. Damaged is an ambiguous, boundary-defined category: its pre-event optical appearance resembles intact, and its SAR response overlaps both neighbors.
 
\par On the cross-event test set, however, the class ranking reverses (Fig.~\ref{fig:test_accuracy}-(b)). Damaged is the strongest class for every leading team, with APs of roughly 0.24 to 0.34, while intact stays near 0.1 and destroyed varies most strongly across teams; the public baseline shows the opposite pattern, 0.055 for intact and near zero for both damage classes.The reversal is informative. Both test events are dominated by affected buildings, so per-class AP reflects the class composition of the event and the calibration of the model to it as much as any intrinsic property of the class. 
 
\par The takeaway is that severity classification has no fixed easy or hard class. Uncalibrated systems collapse on the damage classes, as the baseline did; adapted systems can trade intact accuracy for damage-class accuracy, as the winners did. Progress most likely requires treating severity as an ordinal or explicitly calibrated quantity, with separate attention to the intact--damaged and damaged--destroyed boundaries and to how both move across events, rather than as a third nominal class inside a standard detection loss.

\subsection{Adapting to the Damage Profile of a New Event}
\par The two test events differ sharply in damage profile: in the wildfire scene most affected buildings are destroyed, whereas in the hurricane scene most are damaged but standing (Fig.~\ref{fig:BRIGHT_overview}-(d)). A model trained on a fixed mixture of past events is thus mis-calibrated for both events at once. The distinguishing feature of the first-place solution is that it addressed this directly, classifying the disaster scene, re-calibrating class thresholds per scene type, and self-training on high-confidence, disaster-consistent pseudo-labels drawn from the unlabeled test imagery; these steps, together with model fusion, lifted its test mAP from 0.1095 to 0.1815. The per-event results in Fig.~\ref{fig:per_test_event} make the effect measurable: the first-place solution was the only ranked entry to exceed 0.1 mAP on the destroyed-dominant wildfire, while every other leading team collapsed there and performed adequately only on the damaged-dominant hurricane. This confirms that unlabeled target-event imagery, which is always available in a real deployment, carries usable information about the event's damage profile. Test-time adaptation, event-level prior estimation, and threshold re-calibration should therefore be considered legitimate components of an operational pipeline rather than competition tricks~\cite{wang2021tent}, although pseudo-label self-training carries a known risk of confirmation bias, that is, the reinforcement of the model's own errors~\cite{arazo2020pseudo}, which the present results do not quantify.

\subsection{Metric Design and Reporting}
\par The challenge also suggested that a single mAP number is an imperfect measure of progress on this task. Because the ranking metric averages class-wise APs into one scalar, it is sensitive to how predictions are distributed across the severity classes and to the class composition of the test events, not only to the quality of the underlying model, and it hides the distinct failure modes documented above, in which localization quality, severity accuracy, and event-level calibration evolve differently. Future editions of the challenge should therefore consider refined evaluation protocols, for example scoring building localization and severity classification separately, constraining each predicted geometry to a single severity label, and rewarding calibrated class confidence. In the same spirit, we recommend that work on this task reports per-class AP, per-event scores, footprint recall, and severity calibration as separate diagnostics alongside mAP, so that geometric and semantic progress can be tracked independently.
 
\subsection{Limitations and Outlook}
\par A few limitations should be kept in mind when reading these results. The cross-event test uses only two events, so the transfer findings are indicative rather than statistically firm, and the two events also differ in sensor, season, and geography, so the holdout-to-test gap mixes the effect of a new event with the effect of changed imaging conditions. All image pairs are also supplied co-registered: the challenge deliberately isolates damage mapping from the cross-sensor alignment problem, whereas in operational rapid mapping the archived pre-event optical image and the newly tasked post-event SAR acquisition must first be aligned across different viewing geometries and terrain-induced distortions. At sub-meter building scale, a registration error of a few meters is comparable to an entire footprint, and automated optical--SAR alignment itself remains an open problem with residual errors at the multi-meter level~\cite{chen2026earthobservationdisastermapping, chen2025bright, Yan2026Ultra}. The accuracies reported here therefore presuppose a registration quality that must itself be produced under time pressure.

The winning systems are also engineering-heavy: ensembles of model variants, multi-stage pipelines, and per-scene models increase computational cost, storage, and pipeline management overhead, which matters when time is the scarcest resource after a disaster. The first-place team itself identifies the transition from a scene-segregated to a scene-aware paradigm, in which a single unified network adapts to the disaster type internally, as the natural next step.
 
\par Two further directions stand out. First, both winning solutions built on generic ImageNet-style pretraining for their encoders, whereas no comparable pretrained resource exists for damage evidence in post-event SAR, which every team learned from scratch on \textsc{Bright}; SAR-specific or damage-specific pretraining is therefore a high-leverage direction. Second, with best scores near 0.18 mAP, the task is far from solved. Instance separation in dense building layouts, ordinal severity calibration, and cross-event robustness all leave large headroom. Expanding the benchmark with additional held-out events would make transfer measurements statistically firmer and is planned for future editions of the challenge, together with settings that relax the co-registration assumption or couple alignment and damage mapping in a single task.
 
\section{Conclusion}
\par The \textsc{Bright} Challenge evaluated building damage mapping in the configuration that rapid response actually faces: a pre-event optical image, a post-event SAR image, and the requirement to deliver every building as a detected, delineated, and classified object, scored on disasters absent from the training data. Built on the extended \textsc{Bright} dataset with instance-level annotations, the challenge attracted 157 participants who made 1,289 submissions, with 46 teams active in the final phase. The two winning teams reached test mAPs of 0.182 and 0.181 on the two unseen events, about 8.7 times the public Mask R-CNN baseline, showing that transferable multimodal damage mapping at the instance level is possible.
 
\par At the same time, the challenge exposed how far the task is from solved. Every team scored far lower on the unseen events than on the in-domain holdout, the development-phase leaders were not the final winners, and the relative accuracy of the three severity classes reversed between the two phases, underlining how strongly severity classification depends on event-level calibration. The solutions that transferred best shared a common recipe: modality-specific encoding with staged optical-SAR fusion, decoupled building localization and damage classification, and explicit adaptation to the damage profile of the target event. We believe these findings, together with the evaluation lessons discussed in Section~\ref{sec:takeaways}, offer a concrete starting point for researchers taking up instance-level damage mapping.

\par All challenge resources, including the multimodal imagery, the instance-level annotations, the baseline with training code and weights, and the winning teams' solutions, remain publicly available. We hope they serve as a common reference for developing and, more importantly, stress-testing the next generation of all-weather damage mapping methods.


%



\section*{Acknowledgment}
\par This work was supported by the Council for Science, Technology and Innovation (CSTI), the Cross-ministerial Strategic Innovation Promotion Program (SIP), Development of a Resilient Smart Network System against Natural Disasters (Funding agency: NIED); the JSPS, KAKENHI under Grant Number 24KJ0652, 25K03145, 26K21262, and 26K21244; JST, FOREST under Grant Number JPMJFR206S; Next Generation AI Research Center of The University of Tokyo and RIKEN Incentive Research 2026. Research was sponsored by the Department of the Air Force Artificial Intelligence Accelerator and was accomplished under Cooperative Agreement Number FA8750-19-2-1000. The views and conclusions contained in this document are those of the authors and should not be interpreted as representing the official policies, either expressed or implied, of the Department of the Air Force or the U.S. Government. The U.S. Government is authorized to reproduce and distribute reprints for Government purposes notwithstanding any copyright notation herein.

\par The authors would also like to give special thanks to Sarah Preston of Capella Space, Capella Space’s Open Data Gallery, Maxar Open Data Program, and Umbra’s Open Data Program for providing the valuable data.

\ifCLASSOPTIONcaptionsoff
  \newpage
\fi



\bibliographystyle{IEEEtran}
\bibliography{ref}
\end{document}